\documentclass[preprint]{vgtc}               

\graphicspath{{figures/}{pictures/}{images/}{./}} 

\usepackage{times}                     
\usepackage{tabu}                      
\usepackage{booktabs}                  
\usepackage{lipsum}                    
\usepackage{mwe}                       
\usepackage{mathptmx}                  
\usepackage{float}
\usepackage{amsmath,amssymb}
\newcommand{\tp}[1]{[#1]^\intercal}
\usepackage{siunitx}

\usepackage{enumitem}

\onlineid{0000}
\vgtccategory{Research}
\vgtcinsertpkg

\title{IntelliCap: Intelligent Guidance for Consistent View Sampling}

\author{
    Ayaka Yasunaga $^1$\thanks{e-mail: ayaka.yasunaga@keio.jp}~~
    Hideo Saito $^1$~~
    Dieter Schmalstieg $^2$~~
    Shohei Mori $^{2,1}$\thanks{e-mail: s.mori.jp@ieee.org}}
\affiliation{\scriptsize $^1$ Keio University~~~~$^2$ University of Stuttgart}

\teaser{
  \centering
  \includegraphics[width=\linewidth]{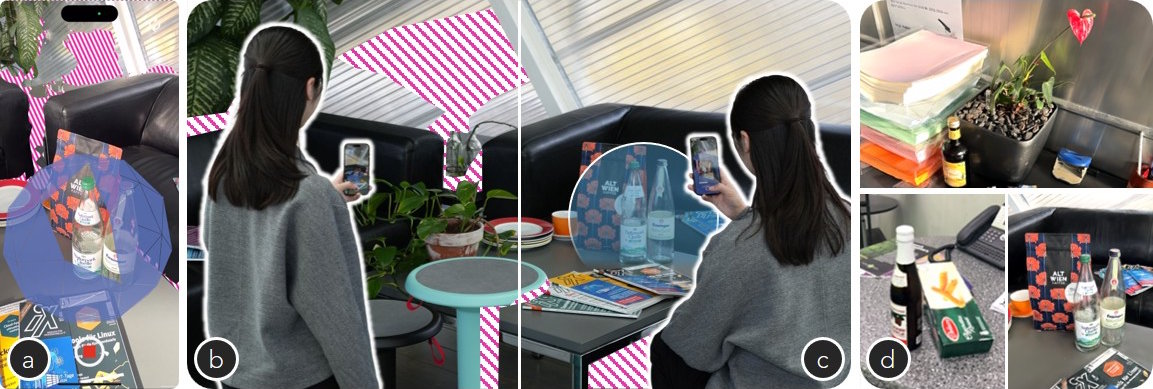}
  \caption{Intelligent guidance for consistent view sampling using a combination of AR and AI tools.
  (a) Our system visualizes areas that lack view samples for spatial and angular coverage to synthesize view-dependent effects (the pink-white stripes and sphere, respectively).
  (b) The user is encouraged to erase the pink-white stripes by pointing the smartphone camera at the visualizations.
  (c) Meanwhile, our system identifies objects that potentially need denser samples evaluated by LLM, and generates spherical proxies to encourage the operator to take more images around them.
  (d) Our approach can avoid exhaustive exploration, yet guarantee the final rendering quality.
  }
  \label{fig:teaser}
}

\abstract{
Novel view synthesis from images, for example, with 3D Gaussian splatting, has made great progress. Rendering fidelity and speed are now ready even for demanding virtual reality applications. However, the problem of assisting humans in collecting the input images for these rendering algorithms has received much less attention. High-quality view synthesis requires uniform and dense view sampling. Unfortunately, these requirements are not easily addressed by human camera operators, who are in a hurry, impatient, or lack understanding of the scene structure and the photographic process. Existing approaches to guide humans during image acquisition concentrate on single objects or neglect view-dependent material characteristics. We propose a novel situated visualization technique for scanning at multiple scales. During the scanning of a scene, our method identifies important objects that need extended image coverage to properly represent view-dependent appearance. To this end, we leverage semantic segmentation and category identification, ranked by a vision-language model. Spherical proxies are generated around highly ranked objects to guide the user during scanning. Our results show superior performance in real scenes compared to conventional view sampling strategies.
} 


\begin{document}
\firstsection{Introduction}

\maketitle

\begin{table*}[t]
    \caption{Qualitative comparisons of visual guidances for novel view synthesis.}\label{tab:comparison}
    \centering
    \footnotesize{
    \begin{tabular}{lllll}
    \toprule
        Method      & Operator & Supported scene scale                    & View-dep. sampling and rendering & Visual guidance            \\
    \midrule
        ActiveSplat \cite{li2024activesplat} & Robot   & Room (2D locations with 360 images)                & No (SplaTAM \cite{keetha2024splatam})                      & None                     \\
        ULF \cite{davis2012unstructured}         & Human   & Single object                  & Yes (ULF \cite{davis2012unstructured})                     & 3D sphere                \\
        MRLF \cite{mohr2020mrlf}        & Human   & Single object                  & Yes (ULF \cite{davis2012unstructured})                    & 3D sphere                \\
        LLFF \cite{mildenhall2019llff}        & Human   & Forward facing scenes          & Yes (MPI \cite{mildenhall2019llff})                    & 3D indicators on 2D grid \\
        FS2MPI \cite{Ishikawa2023FS2MPI}      & Human   & Forward facing scenes          & Yes (MPI \cite{mildenhall2019llff})                     & 2D rectangle \\
        Ours        & Human   & 3D open-space with multiple objects               & Yes (3DGS \cite{Kerbl20233DGS} and NeRF \cite{nerfstudio})                     & Progressive 3D scene mesh and spheres  \\
        \bottomrule
    \end{tabular}
    }
\end{table*}

New methods for novel view synthesis from images have made great progress. For example, 3D Gaussian Splatting (3DGS) and neural radiance fields (NeRF) synthesize photorealistic images at high frame rates \cite{Kerbl20233DGS,mueller2022instant}. The underlying scene representations are optimized using differentiable rendering. A key requirement for differentiation is uniform and dense sampling of the input views, so that a reliable estimation of gradients can be performed. 

Unfortunately, human camera operators often lack the skill or motivation to acquire such uniform and dense samples. Time limitations may prevent enough images from being collected. Individuals with limited knowledge of view synthesis may not understand the required strategy for image taking. After a lengthy scene optimization process, which can take tens of minutes to hours, it may not be possible to continue image acquisition. The severity grows with the size of the scene that should be acquired. While existing strategies work well for scanning individual objects \cite{davis2012unstructured,mohr2020mrlf}, solutions that scale to larger scenes or open areas are lacking \cite{mildenhall2019llff}.

Instant visual guidance using various forms of augmented reality (AR) has been proposed to assist individuals in image acquisition. The guidance tool can visually indicate where to take more photos or show intermediate results of the ongoing scanning and reconstruction progress. For example, 3D annotations are popular for ensuring aliasing-free light field rendering in the near field based on plenoptic sampling theory~\cite{mildenhall2019llff}, but do not trivially scale to larger areas. Volumetric reconstruction~\cite{newcombe2011kinectfusion} and 3DGS SLAM ~\cite{matsuki2024gaussian} show which parts of the scene are already covered. However, they are only able to cover large areas by trading off resolution and view-dependent properties for robustness and performance.

We aim to provide a unified solution for guiding image acquisition that works on multiple scales and supports complex scenes without sacrificing quality (\figurename~\ref{fig:teaser}). We automatically identify objects and determine the required amount of image coverage matching the object's topology and material properties. We rely on AI tools to segment objects and estimate their visual complexity, such as topology, reflective, and diffractive materials. A vision model, Detectron2 \cite{wu2019detectron2}, performs object segmentation and semantic categorization of objects. Semantic categories are ranked using a large language model (LLM) to inform the image-taking process. 

With this information, spherical indicators are shown in AR to support the user's sampling procedure. We adapt the AR visualization to the object scales and scene geometry to avoid suggesting unreachable areas. To enable operation on mobile devices, our system can offload online vision recognition tasks from mobile devices to a server by transmitting selected keyframes from the view samples.

We validate our approach by comparing it with a conventional guidance approach that determines occupancy and information gain of a 3D reconstruction. To evaluate our work, we develop an evaluation scheme for this new task to compare multiple 3DGS datasets collected by different participants in open environments.

The contributions of this work can be summarized as follows:
\begin{itemize}
\item 
We propose an AR guidance system for quality view synthesis that uses semantic categorization of objects seen in input views and LLM processing to determine scanning requirements of observed objects.
\item 
We present two situated visualizations in combination, consisting of a scene (e.g., temporal 3D reconstruction) and objects abstracted as spheres, for comprehensive spatial and angular view sampling.
\item 
We demonstrate superior performance in 3DGS and Nerfacto view synthesis using view samples from our approach through a user study and rendering quality assessment. To achieve this, we developed an evaluation scheme to assess this new task of view sampling with AR guidance for view synthesis.
\end{itemize}

\section{Background and related work}\label{sec:related_work}

\begin{figure*}[t]
    \centering
    \includegraphics[width=\textwidth]{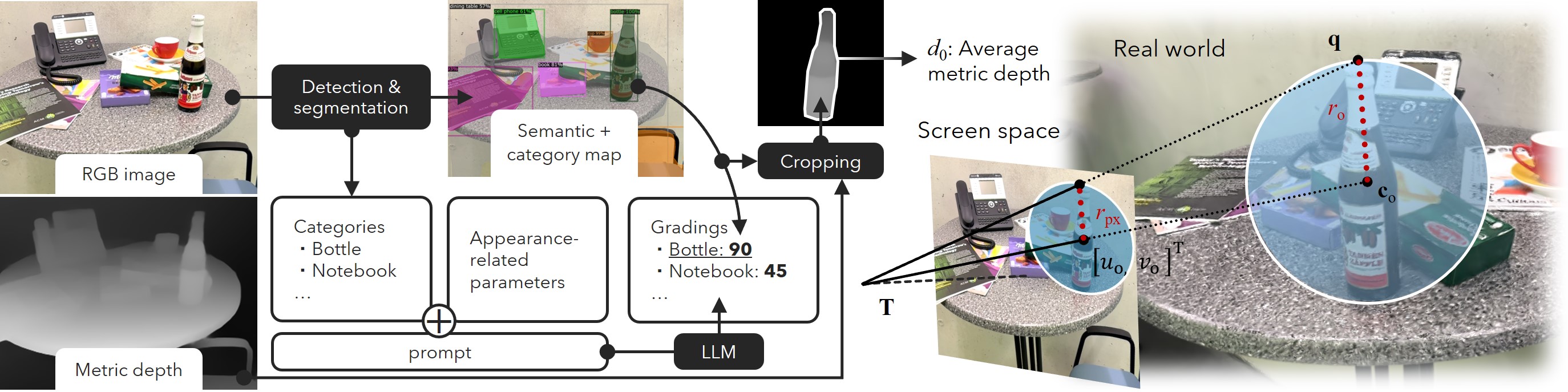}
    \caption{3D sphere generation.
    Given a pair of RGB and metric depth maps, our system detects important objects for view sampling and projects them into real-world space as spheres.
    We use an LLM with a custom prompt template to quantify the visual characteristics of the detected objects and prioritize them accordingly.
    Objects with high scores are anchored in the 3D scene to indicate areas where the user should focus their sampling efforts.
    }
    \label{fig:sphere_generation}
\end{figure*}

We discuss the limitations of current visual guidance approaches for view sampling in open 3D scenes and elaborate on our key ideas to compare them with existing solutions in related areas.

\subsection{Background}

The challenge of visual guidance for view sampling lies in indicating where to sample views without having access to the final view synthesis results. \tablename~\ref{tab:comparison} summarizes the current approaches.

\paragraph{Small area coverage}
Current solutions either rely on geometric scanning \cite{li2024activesplat}, which lacks view-dependent sampling, or are based on the well-established plenoptic sampling theory \cite{chai2000plenoptic}. The latter, which ensures anti-aliased view synthesis, applies mainly to forward-facing cameras due to its assumption of a fixed scene depth range \cite{mildenhall2019llff}. However, in real-world open scenes, the minimum and maximum depths vary depending on the scene and the view frustum. Thus, the applicability of the theory is limited.

Earlier 3D scene representations, such as multi-plane images (MPI) \cite{mildenhall2019llff,Ishikawa2023FS2MPI}, require hundreds of images to cover even a square meter of space. Their visual guidance systems focus on precisely aligning the camera with 3D indicators at predefined positions.
In contrast, modern approaches such as 3D Gaussian splatting (3DGS) can cover significantly larger areas with the same number of images, enabling new possibilities for visual guidance in open environments. Despite this improvement, the current best practice remains to capture as many images as possible to conquer scene complexity.

\paragraph{Missing knowledge of scene objects}
While view-dependent effects must be thoroughly captured, data acquisition should be completed in a practical timeframe. Therefore, the sampled areas must be prioritized. For example, glass bottles or leafy plants demand denser sampling than simple planar objects like desks.
Novice users often lack intuition for how many images to take until sampling is sufficient. They lack a visual guidance system that can signal the completion of the sampling and suggest attention to areas of interest. However, scoring individual objects during photo capture remains a challenge.

\paragraph{Our solution}
To tackle these two technical challenges, we propose (1) progressive visual indicators using 3D mesh and sphere proxies, and (2) LLM-based scoring of detected objects and their categories. Users walk freely through the environment, while our system performs 3D reconstruction to evaluate spatial coverage. Meanwhile, the system detects objects and generates spherical proxies to attract user attention (\figurename~\ref{fig:teaser}b).
As users continue scanning, new objects may be found, spawning additional proxies to prompt denser view sampling.
Once all highlighted visualizations disappear, the user can consider the view-sampling process complete.
Alternatively, users may keep expanding the sampled areas by searching for the remaining attractions in the environment.
This is naturally supported by the progressive operation of our system.

\subsection{Visual guidance for view sampling}

AR can visually indicate where to operate a camera and collect image samples \cite{mildenhall2019llff, Ishikawa2023FS2MPI}. Popular forms of annotations include 3D axes \cite{mildenhall2019llff}, 2D planes \cite{Ishikawa2023FS2MPI, Birklbauer2015Active}, and hemispheres \cite{davis2012unstructured, mohr2020mrlf} surrounding a target area. These methods recast the photographing task as a data collection task. For example, when using a smartphone, the 3D annotations can be directly overlaid on the viewfinder display. The user must then navigate the phone to ``intersect'' \cite{mildenhall2019llff, Ishikawa2023FS2MPI, Birklbauer2015Active} it with the 3D annotations or ``cast a ray'' \cite{davis2012unstructured, sun2022onepose, mohr2020mrlf} hitting the annotation. When the geometric constraints indicated to the user are met, the smartphone automatically captures the image and advances to the next annotation. The user only has to collect all images indicated by the annotation, which often changes color to indicate which parts have been fulfilled already.

During capture on the mobile device, the annotations are derived from theoretical bounds such as the required sampling rate. Performing even a preliminary analysis-by-synthesis to steer the feedback would be too costly in these circumstances. Again, typical approaches place 3D annotations at the locations according to the plenoptic sampling theory \cite{chai2000plenoptic, ng2005fourier}. The locations are fixed at the beginning of the photo session. This process involves a detailed scan of the scene before starting the guidance, so sampling rates can be guaranteed \cite{mildenhall2019llff, Ishikawa2023FS2MPI, Birklbauer2015Active}. This process is rather rigid and is not suitable for spontaneous capture sessions. Existing approaches aim at capturing only one target object at a time \cite{davis2012unstructured, Birklbauer2015Active} or a small area in front of the operator \cite{mildenhall2019llff, Ishikawa2023FS2MPI, li2024activesplat}.

Our approach dynamically evolves the capture area by detecting unobserved regions and newly discovered objects that require denser sampling. It combines standard 3D scanning techniques with spherical proxies to provide intuitive visual cues suited for open scenes, while leveraging computer vision to automate scene understanding. This enables adaptive and progressive visualization, relieving users from the burden of manually assessing scene complexity and coverage.

\subsection{Real-time view synthesis as visual feedback}

Real-time reconstruction can provide immediate visual feedback on rendering quality, allowing users to continue sampling new views until satisfied. The state-of-the-art online reconstruction of neural view synthesis hardly meets real-time requirements due to its costly optimization process \cite{zhu2022nice-slam, Zhu2024NICER}, except when a dedicated view synthesis pipeline is used \cite{fink2023livenvs}.
The latest 3DGS reconstruction methods run at interactive frame rates, but prune details for increased speed and robustness \cite{MuraiCVPR2024, keetha2024splatam}. For full-quality reconstruction, batch optimization must be performed later on \cite{keetha2024splatam}.

A substitute for the full reconstruction used in newer commercial applications relies on real-time volumetric reconstruction (cf. KinectFusion \cite{newcombe2011kinectfusion} and DTAM \cite{newcombe2011DTAM}). However, these volumetric reconstructions use a coarse resolution and are mostly used to estimate scene coverage and occlusion, but they cannot ensure that sufficient geometric or photometric details are captured.

To compensate for the missing fine detail during capture, we introduce spherical proxies that indicate areas that require additional angular view samples. We implement this approach on a smartphone and demonstrate its effectiveness compared to standard methods that highlight missing regions in real-time reconstruction.

\subsection{Next best view}

Research on uncertainty analysis in neural rendering and 3DGS has targeted floater removal \cite{goli2024bayes}, compression \cite{HansonTuPUP3DGS} and decision making about the next best view \cite{jin2023neunbv}. To find the next best view samples online, geometric information gain is evaluated \cite{li2024activesplat} by rendering 360 views at candidate locations and selecting the most empty view. Next-best-view finding involves on-demand scene optimization, which is affected by the same computational limits as applied to SLAM systems. Reported results rely on simulated environments assuming a pre-scanned mesh \cite{erat2019real}, hours of intermediate training \cite{sunderhauf2023density}, or systematic dense uniform scanning \cite{Guthe2023Progressive}, all of which is not applicable for real-time scanning sessions operated by humans.

Our system is designed for fully online operation, offloading computer vision processes to a server and utilizing pre-computed object importance priors provided by an LLM. The smartphone is primarily responsible for 3D scanning, visualization, and photo capture. This lightweight configuration enabled our user study. While our system does not provide explicit navigation to the next-best-view locations, it allows users to identify them through intuitive visualizations.

\section{Method}

We developed a system that captures images at fixed time intervals as the user walks through the environment.
To guide the user, the system provides two types of visual highlights in the 3D scene: one based on spatial coverage, and another based on angular coverage for objects that may impact view synthesis quality due to their photometric complexity (\figurename~\ref{fig:teaser}a).
These visualizations are progressively updated and refined as view sampling progresses (\figurename~\ref{fig:teaser}b). The collected view samples are then used to optimize 3DGS for rendering the corresponding scenes (\figurename~\ref{fig:teaser}c).

\subsection{Spatial and angle coverage analysis}
\label{sec:scene_analysis}

\paragraph{3D mesh proxy for spatial coverage}
Existing methods estimate scene coverage \cite{li2024activesplat, sucar2021imap, zhu2022nice-slam} by constructing a 3D point cloud.
Surfaces with sparse or missing points are identified as under-sampled and targeted for densification.
In our approach, we highlight regions that have not yet been reconstructed, based on the current state of the 3D reconstruction (\figurename~\ref{fig:teaser}a). Incomplete areas are overlaid with pink-white stripes, while the rest is taken from the live camera feed.
This provides visual feedback about which areas have already been visited or not. For simplicity and performance in implementation, we utilize the smartphone's native scene meshing API (ARCore or ARKit).

\begin{figure*}[t]
    \centering
    \includegraphics[width=\textwidth]{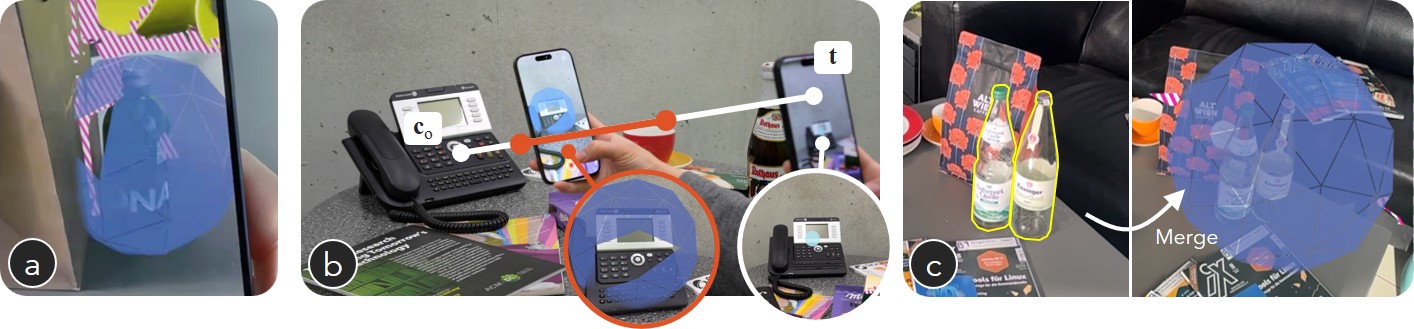}
    \caption{Scene adaptive 3D sphere visualization.
    (a) Scene-adaptive rendering guides the user and prevents them from walking into unreachable areas.
    (b) Spheres are only displayed within a valid depth range illustrated in orange to avoid sampling from positions too close or too far.
    (c) Two nearby spheres are merged into a single sphere, inheriting the properties of the older one. This naturally supports the integration of multi-view inputs for the same object from different viewpoints.
    }
    \label{fig:vis_adaptive}
\end{figure*}
\begin{figure}[!b]
    \centering
    \includegraphics[width=\columnwidth]{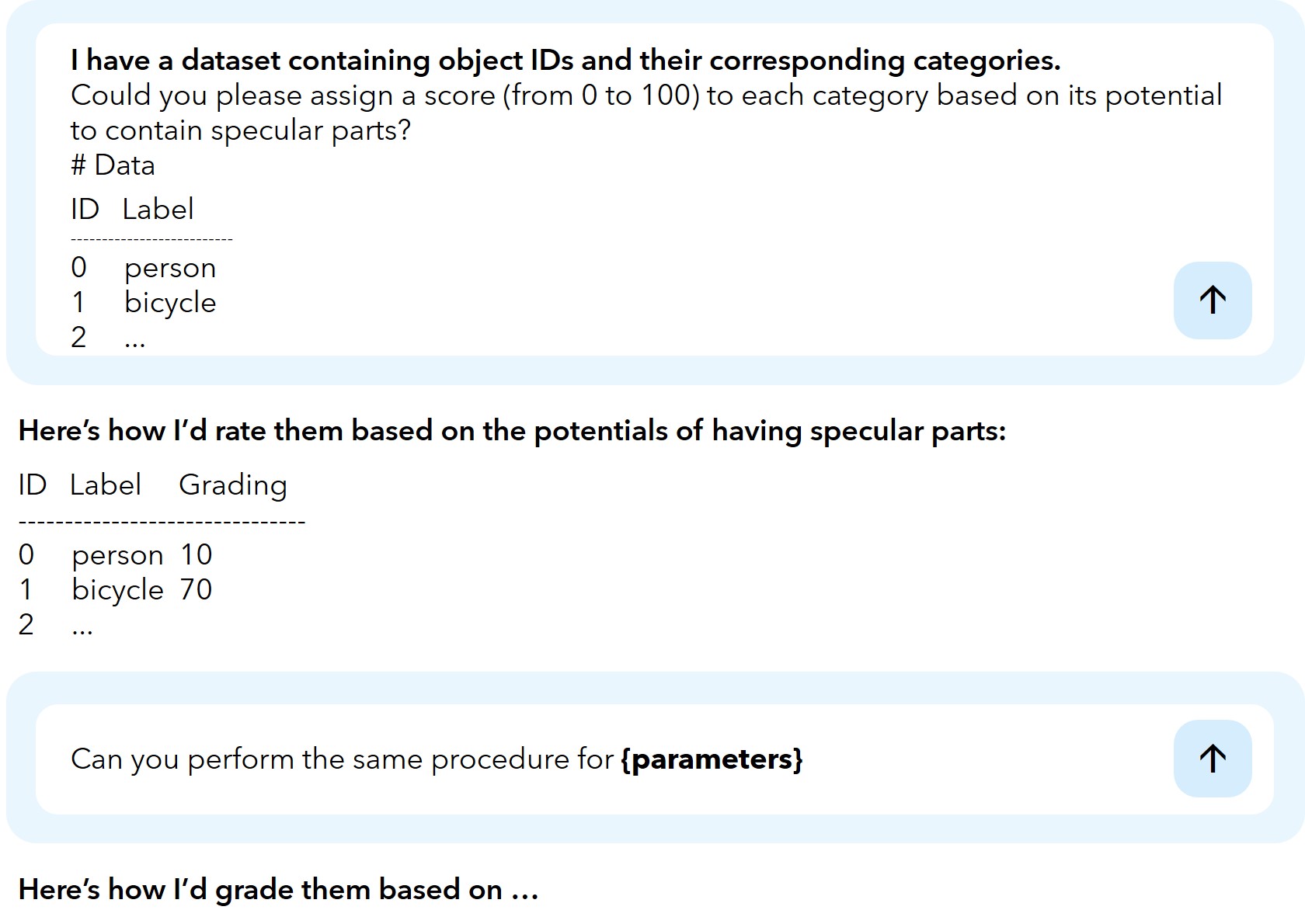}
    \caption{The prompt for scoring the object categories to quantify their appearance characteristics.}
    \label{fig:copilot}
\end{figure}

\paragraph{3D sphere proxy for angular coverage}
While the mesh proxy supports spatial coverage, it does not account for angular sampling. Users may miss critical photometric details, such as specular reflections or transparency, that vary significantly with viewpoint and are known to be challenging to reconstruct without sufficient angular diversity \cite{Guthe2023Progressive}.
To address this, we detect objects likely to exhibit such photometric complexity during the walkthrough. Rather than identifying all objects that can easily occupy the scene with spheres, we selectively detect only those that benefit most from denser view sampling. Only for these, we generate 3D sphere proxies \cite{davis2012unstructured, mohr2020mrlf}.

Each sphere proxy consists of a 3D position $\mathbf{c}$, a radius $r$, a set of subsurface $S \in \{s_i | 0 \leq i \leq N\}$, as $(\mathbf{c}, r, S)$. Each subsurface corresponds to a specific view sample, allowing us to control the sampling density and coverage speed. As users capture new views, the corresponding subsurfaces are marked as covered and disabled by setting their transparency to full. The subsurfaces are retained throughout the session.
The marked subsurface becomes transparent to tell the capture to complete as in \figurename~\ref{fig:teaser}a.

In the following sections, we describe how these 3D spheres are generated and how their representations are dynamically updated.

\subsection{3D sphere generation}

\figurename~\ref{fig:sphere_generation} provides an overview of the 3D sphere generation process.
The input to our guidance system consists of multiple captured views, each comprising an RGB image, a metric depth map, and intrinsic and extrinsic camera parameters. These input views are processed independently.
We use Detectron2~\cite{wu2019detectron2} to detect object segments and their categories from RGB images. The depth maps are used to estimate object size in the 3D scene. Identified objects are then analyzed to determine whether they should be registered for denser view sampling.
Since it is typically unclear from a single view how geometrically or photometrically complex an object is, we use object semantics to bridge this gap.

The semantics in the form of object categories are processed with an LLM to determine an estimate of the visual object complexity.
The key idea is to use a foundation model, which contains common human knowledge, to obtain properties that are difficult to quantify.
Therefore, we preferred a closed vocabulary to provide the necessary knowledge efficiently.
We create a prompt template and recursively ask to score the scores for objects' appearance characteristics.
The template prompt is defined as follows, `\texttt{I have a dataset containing object IDs and their corresponding categories. Could you please assign a score (from 0 to 100) to each category based on its potential to contain \{parameter\}? \{object list\}},' where \texttt{\{parameter\}} $\in$ \texttt{\{geometric complexity, texture complexity, size, specularity, transparency\}}. The questions for the rest of the parameters in a sequence to keep the context (\figurename~\ref{fig:copilot}).
The category labels are pre-scored by an LLM (e.g., MS Copilot\footnote{\url{https://copilot.microsoft.com/}}).
The scores of all metrics are averaged for the final score.
Once an object is found that exceeds a threshold, the object is tagged for higher angular sampling.

The tagged objects are approximated with 3D spheres. The sphere visualizes the coverage, indicating from which viewing angles the object was already observed \cite{davis2012unstructured, mohr2020mrlf}. In our approach, subsurfaces are evenly distributed on the sphere for uniform capture \cite{saff1997distributing}.
The spheres are anchored to the real objects.
We assume that the center of the sphere is at the centroid of the segmented object in screen space $\tp{u_\text{o},v_\text{o}}$ with associated depth coordinate $d_o$, which is computed as the average depth of the object samples (after outlier removal). We transform the sphere center $\mathbf{c}_\text{o}=\tp{x_\text{o},y_\text{o},z_\text{o}}= \mathbf{R} \mathbf{K}^{-1} d_\text{o}\tp{u_\text{o}, v_\text{o}, 1} + \mathbf{t} $ into world coordinates using the camera position $\mathbf{t}$, rotation matrix $\mathbf{R}$, and intrinsic matrix $\mathbf{K}$.
The radius of the sphere in pixels $r_\text{px} = k\max(u_{\text{max}} - u_{\text{min}}, v_{\text{max}} - v_{\text{min}})$ is calculated from the 2D bounding box of the object, which has the corners $\tp{u_{\text{min}}, u_{\text{max}}}$ and $\tp{v_{\text{min}}, v_{\text{max}}}$. A scaling factor $k$ ensures that the sphere fully encloses the object. To convert the radius into metric units, we assume an auxiliary point $\mathbf{q}=\mathbf{R} \mathbf{K}^{-1} d_\text{o}\tp{u_\text{o},v_\text{o}-r_\text{px},1} + \mathbf{t}$ located above the sphere center. The metric radius of the new sphere is determined with the help of $\mathbf{q}$ as $r_\text{o}=||\textbf{q}-\textbf{c}||$.

\subsection{Dynamically adjusting 3D spheres}

\paragraph{Occlusion handling}
To prevent confusion about occluded areas, we avoid overlaps between spheres and scene objects, helping users understand that they do not need to capture views behind occluders such as walls.
The alpha value (transparency) of each sphere is dynamically adjusted based on its depth relationship with surrounding scene elements. Specifically, if the depth of the sphere is closer than the depth of the scene at a given pixel, the sphere is displayed. Otherwise, it is hidden behind the scene geometry.

To mitigate potential misalignment caused by the imperfect scene mesh under reconstruction, we use real-time depth streams from the smartphone. In addition, we render soft edges at the intersections between spheres and scene geometry to provide visual tolerance.
The level of sphere transparency is determined by the offset in depth $\Delta d / t$, where $\Delta d$ is the depth difference between the sphere surface and the scene, and $t$ is a tolerance threshold set to 5.0 cm by default (\figurename~\ref{fig:vis_adaptive}a).

\begin{figure*}[t]
    \centering
    \includegraphics[width=\textwidth]{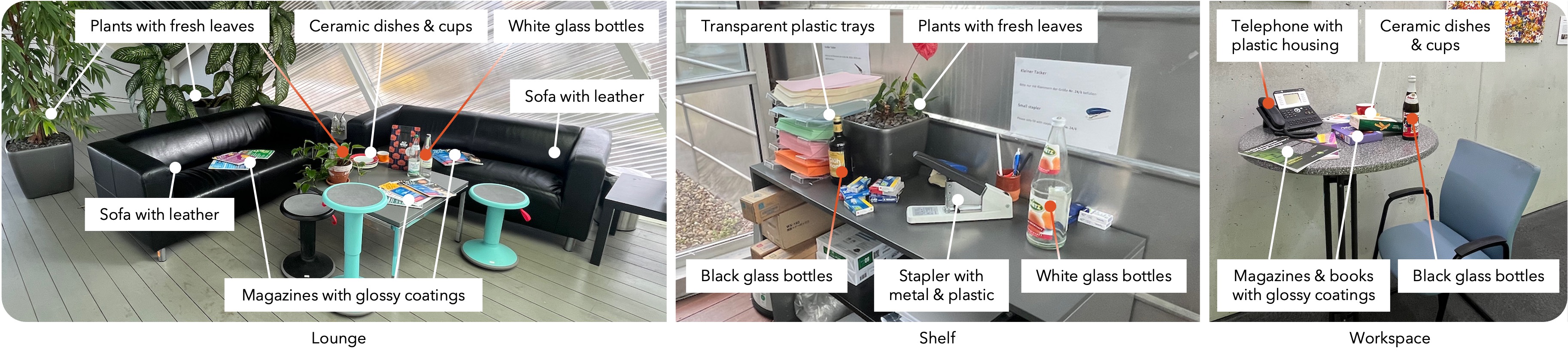}
    \caption{Large ($\approx$ 9 m$^2 = 3 \times 3$), medium ($\approx$ 2 m$^2 = 2 \times 1$), and small ($\approx$ 1 m$^2 = 1 \times 1$) scenes for the experiments.
    To increase visual complexity, we intentionally arranged white and black glass bottles, plastic bottles, ceramic cups and dishes, plants with fresh leaves, books and magazines with glossy coatings, and a telephone with plastic housing. The objects highlighted with red lines have high potential to be abstracted as spherical proxies if they are appropriately detected as one of the complex objects of labels: \texttt{vase}, \texttt{bottle}, \texttt{cellphone}, etc.}
    \label{fig:exp_setup}
\end{figure*}

\paragraph{Distance-based suppression}
To keep the object of interest within the user’s field of view (FoV), spheres are visualized only when they fall within a valid distance range (\figurename~\ref{fig:vis_adaptive}b). When outside this range, a fixed-size dot is shown instead, indicating the location of an object identified by the system.
View sampling cannot be completed unless the user approaches the sphere within the required distance, naturally encouraging closer interaction.
We allow the full sphere to appear once it occupies more than 20\% of the camera FoV. When it reaches 100\%, the sphere disappears, indicating that the user is too close and that view sampling at this range is unnecessary.

\paragraph{Merging spheres}
As the number of spheres increases, the workload for capturing becomes higher. Therefore, our objective is to minimize the number of spheres by intelligently merging expendable spheres (\figurename~\ref{fig:vis_adaptive}c). Initially, a collision check is performed between the spheres, and only intersecting spheres are considered for merging. In the trivial case where a sphere is completely inside another sphere, the smaller sphere can simply be discarded. Otherwise, the two spheres $(\mathbf{c}_1,r_1,S_1)$ and $(\mathbf{c}_2,r_2,S_2)$ are merged into a new sphere $(\mathbf{c}_\textbf{new}, r_\text{new}, S_\text{new})$. We compute the center of the new sphere as the midpoint between the two sphere centers, $\mathbf{c}_\text{new}=(\mathbf{c}_1+\mathbf{c}_2)/2$ and the radius as the sum of the radii plus the distance between the sphere centers, $r_{\text{new}} = (r_1 + r_2 + ||\mathbf{c}_1-\mathbf{c}_2||)/2$. The radius of the new sphere is truncated at a maximum value to avoid creating excessively large spheres.
This cap also prevents generating multiple spheres at nearly the same position for a single object through multi-view input.
The new sphere succeeds the view sample coverage from the old sphere by $S_\text{new} = S_1 \cap S_2$.

\section{Evaluation}

We evaluate our system in terms of usability, task load, and the resultant view synthesis quality from images collected by our approach at different scene scales.

\subsection{Study setup}

\paragraph{Design}
We designed a repeated-measures within-subjects study to identify the characteristics of our proposed visualization approaches.
We compared the following three approaches (i.e., independent variables in this study):
\begin{itemize}[left=0pt, itemsep=0pt, parsep=0pt, partopsep=0pt]
    \item \texttt{NV}: The video stream is presented without any guidance.
    \item \texttt{SC}: Only spatial coverage is visualized. This is analogous to approaches with spatial analysis.
    \item \texttt{Ours}: Both spatial and angular coverages are visualized.
\end{itemize}

\paragraph{Metrics}
To evaluate the system implementations, we collected System Usability Scale (SUS) \cite{lewis2018system}, NASA Task Load Index (NASA-TLX) \cite{hart1988development}, Satisfaction (Satis.), and the number of collected images. Satisfaction measured evaluations on a 7-point scale between (1) ``Totally disagree'' to (7) ``Totally agree,'' with the question: ``I am satisfied with the scene scanning.'' per method.
Our post-task questionnaires were designed specifically around the view sampling task. Participants ranked the three approaches for each of the following questions:
(Q1): ``Which interface did you find most comfortable?'' (comfort),
(Q2): ``Which interface did you find the most enjoyable to use?'' (enjoyment),
(Q3): ``Which interface most effectively supported your task?'' (task support),
(Q4): ``Which interface felt the most intuitive and natural to use?'' (naturalness),
(Q5): ``Which interface provided a better sense of spatial awareness?'' (spatial awareness), and
(Q6): ``Which interface did you prefer overall?'' (overall performance).
We also collected open-ended feedback by asking participants to comment on the following questions: ``What was the hardest part about performing the tasks?'' and ``What do you think can be improved in general?''.
For evaluating view synthesis quality with the collected image data, we processed the multi-view images and generated Nerfacto \cite{nerfstudio} and 3DGS \cite{Kerbl20233DGS} scenes.
We measured image quality using standard metrics, including peak signal-to-noise ratio (PSNR), structural similarity (SSIM) \cite{wang2004image}, and LPIPS \cite{zhang2018unreasonable}.

\paragraph{Participants}
We collected 12 participants (two female and 10 male) with an average age of $\Bar{X}=29.2$ (SD$=4.2$) years, all right-handed and with corrected vision. All participants were university students in computer science and scored their AR experience as $\Bar{X}=5.2$ (SD$=1.5$) on a 7-point scale, where 1 represents `never experienced' and 7 represents `regular user.'
This experiment was approved by the local institutional review board.

\paragraph{Apparatus}
We used the Apple iPhone 15 Pro. The software was implemented using Unity, C\#, and Shader Lab, with Unity ARFoundation for augmented reality features, including device tracking and per-frame depth maps. The captured images are saved asynchronously on the local device every $0.2$ seconds, and keyframes are sent to the server every $5$ seconds for vision processing.
On the server side, we use a desktop PC with an AMD Ryzen 9 7900X 4.7 GHz CPU, 64 GB RAM, and an NVIDIA GeForce RTX 4090 24 GB GPU. Data is communicated between the mobile device and the server using Hypertext Transfer Protocol (HTTP).

\paragraph{Procedure}
After filling out a consent form, each participant was introduced to a brief training session, which took approximately five minutes. They were taught to hold and move the device in portrait mode. They learned the purpose of view sampling and the way to complete view sampling tasks with each visualization approach.
They were instructed to complete the task as quickly and precisely as possible.
After the training session, the main session commenced.
Participants were invited to one of the three environments (\figurename~\ref{fig:exp_setup}), where one of the visualization approaches was selected and presented. To avoid biases, we employed a $3\times3$ Latin square design.
Participants began the scene capturing by taking an initial image and continued until they felt the sampling was sufficient.
After the view sampling session, they answered questionnaires. This procedure was repeated until all visualizations were evaluated. Finally, participants were asked follow-up questions. The entire procedure took approximately one hour.

\paragraph{Dataset}

We conducted our user study in three scenes (\figurename~\ref{fig:exp_setup}). The design of our dataset is analogous to the Shiny dataset \cite{wizadwongsa2021nex}, which primarily evaluates view-dependent effects using MPI and spherical harmonics-based view synthesis. The Shiny dataset consists of scenes containing specular objects. Similarly, we arranged objects with reflective and transparent materials to emphasize challenging view-dependent phenomena.

In most view synthesis research, a single person collects all view samples for a scene, and these are split into training and evaluation sets.
This approach, however, only evaluates performance on views close to the original sampling trajectory and is not suitable for comparing different view sampling strategies, which is our focus.
In fact, there is a report that biased ground truth selection can strongly influence final view synthesis results \cite{Xiao:CVPR24:NeRFDirector}.
To address this, we propose using an independent set of ground truth captured by another person (i.e., an examiner) to exhibit broader spatial and angular variance.
Because the scenes were open environments with naturally changing lighting conditions, an examiner collected the ground-truth image set immediately after each participant’s trial to minimize scene drift between training and evaluation datasets.

All input images were captured at 1920$\times$1440 pixels and resized to 480$\times$360 pixels for vision processing.
We used COLMAP \cite{schoenberger2016sfm}, a widely adopted structure-from-motion software, to estimate camera poses for both the participants’ images and 20 ground truth images.
These ground truth images were randomly selected from the full stock to avoid intentional selection and potential biases as much as possible.
Using the participants’ input data, we optimized Nerfacto and 3DGS of their official implementations\footnote{Nerfacto v1.0.1: \url{https://github.com/nerfstudio-project/nerfstudio} and 3DGS (Commit: 54c035f): \url{https://github.com/graphdeco-inria/gaussian-splatting}} and synthesized views at the ground truth viewpoints.

\subsection{View synthesis quality}
We compared our approach with two baseline approaches and quantitatively evaluated how effective our view sampling strategy is in both scene representations.

\begin{figure*}[!t]
    \centering
    \includegraphics[width=\textwidth]{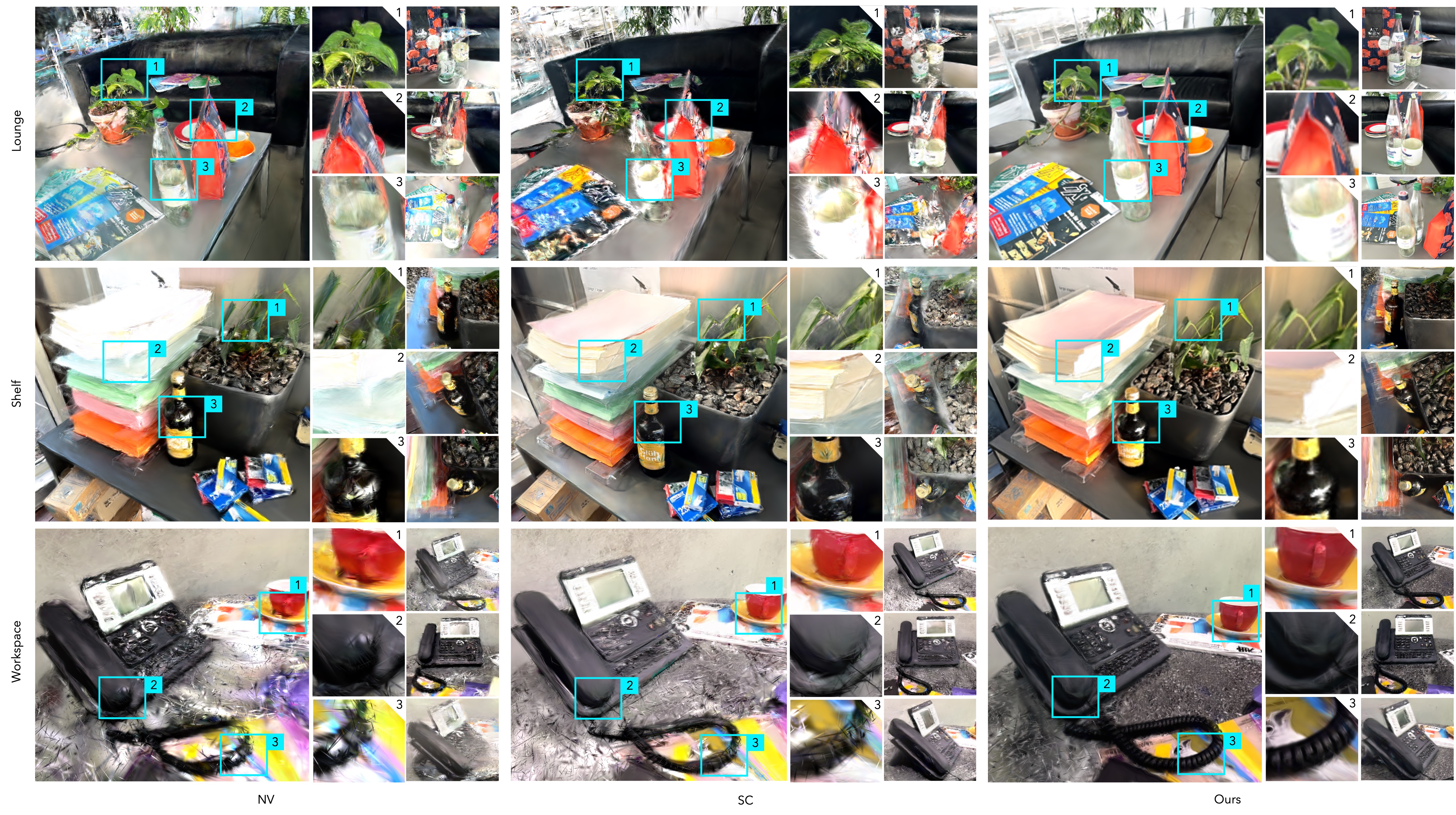}
    \caption{3DGS \cite{Kerbl20233DGS} view synthesis results using data collected by ours and baseline approaches.
    The rightmost columns in each approach show how reflective and transparent objects are reproduced at different angles from the collected views.
    \texttt{Ours} exhibits fewer artifacts across different scenes. \texttt{NV} suffers from artifacts all around the scenes, and \texttt{SC} fails to reproduce detailed and glassy objects.
    }
    \label{fig:3dgs_results}
\end{figure*}

\begin{figure*}[!t]
    \centering
    \includegraphics[width=\textwidth]{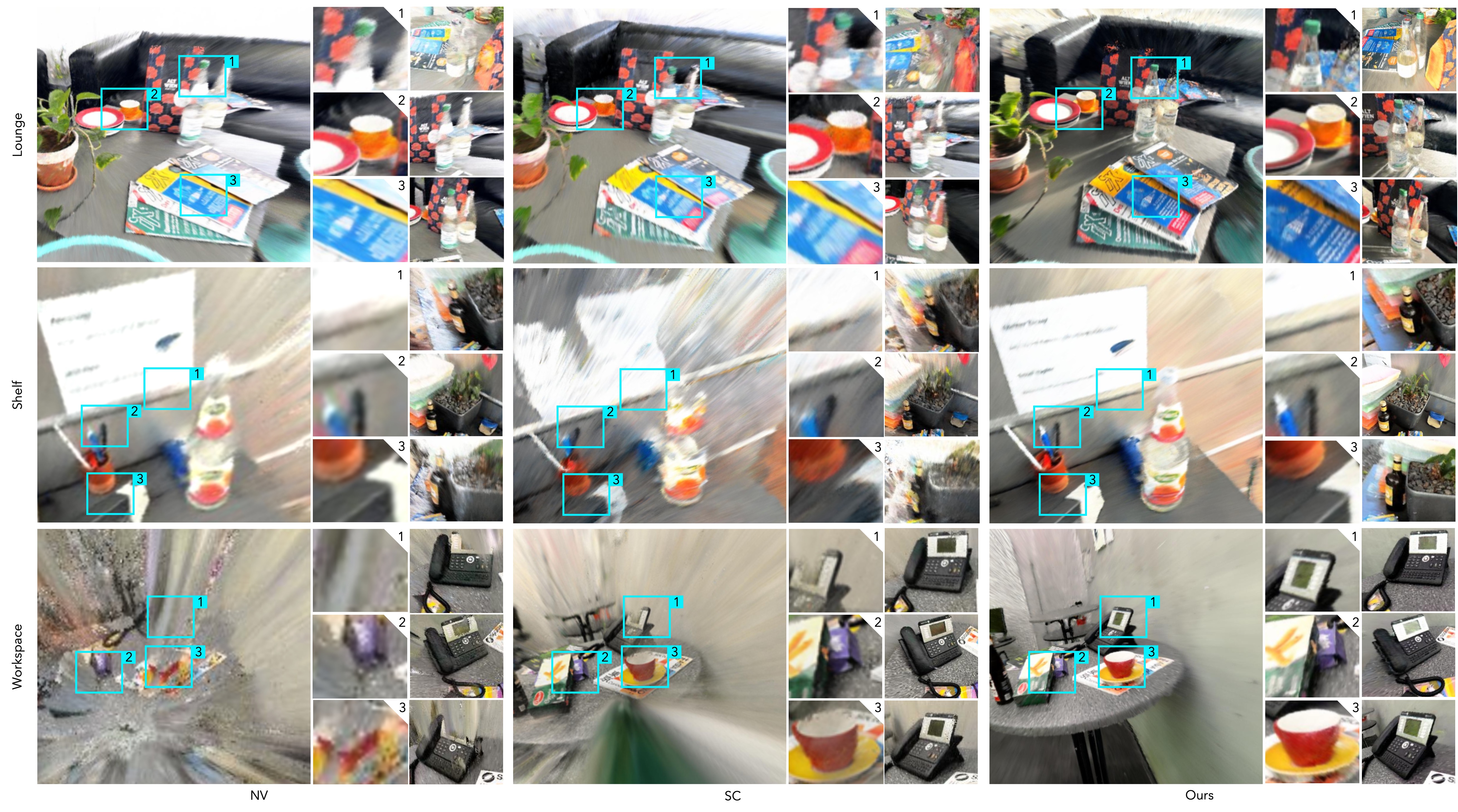}
    \caption{Nerfacto \cite{nerfstudio} view synthesis results using data collected by ours and baseline approaches.
    The rightmost columns in each approach show how reflective and transparent objects are reproduced at different angles from the collected views.
    While \texttt{NV} and \texttt{SC} show characteristic artifacts of neural rendering, \texttt{Ours} provides coherent view synthesis across the test scenes.
    }
    \label{fig:nerf_results}
\end{figure*}

\paragraph{Results}
\tablename~\ref{table:results} summarize the results of the 3DGS and Nerfacto view synthesis.
Our method outperforms the baselines in all image quality metrics.
\texttt{Ours} achieved the highest performance, followed by \texttt{SC} and \texttt{NV}.
Figures~\ref{fig:3dgs_results} and \ref{fig:nerf_results} present qualitative comparisons of 3DGS and Nerfacto view synthesis results among the three methods.
Our approach generates high-quality, consistent reconstructions throughout the entire scene, supported by spatial and angular coverage analysis.
As demonstrated in the object-centered rendering results, our method effectively captures challenging aspects such as specular reflections and transparency, which baseline methods often miss. These view-dependent effects are most clearly illustrated in the supplemental materials.

\paragraph{Discussion}
Overall image quality scores are relatively lower ($<$ 20 dB) compared to those typically reported in the literature (23--30 dB).
As discussed in the previous section, our view synthesis task is more challenging than the standard benchmarks. However, if we follow the standard practice used in the literature and select ground truth images from among participant-captured images, the scores for all three methods improve significantly by 37\%, 15\%, and 49\% in PSNR, SSIM, and LPIPS, respectively.
This improvement reaches an average PSNR of approximately 24 dB, which is in agreement with the expected scores and confirms the difficulty of this new view sampling task and our evaluation setting.

\begin{figure*}[t]
    \centering
    \includegraphics[width=\textwidth]{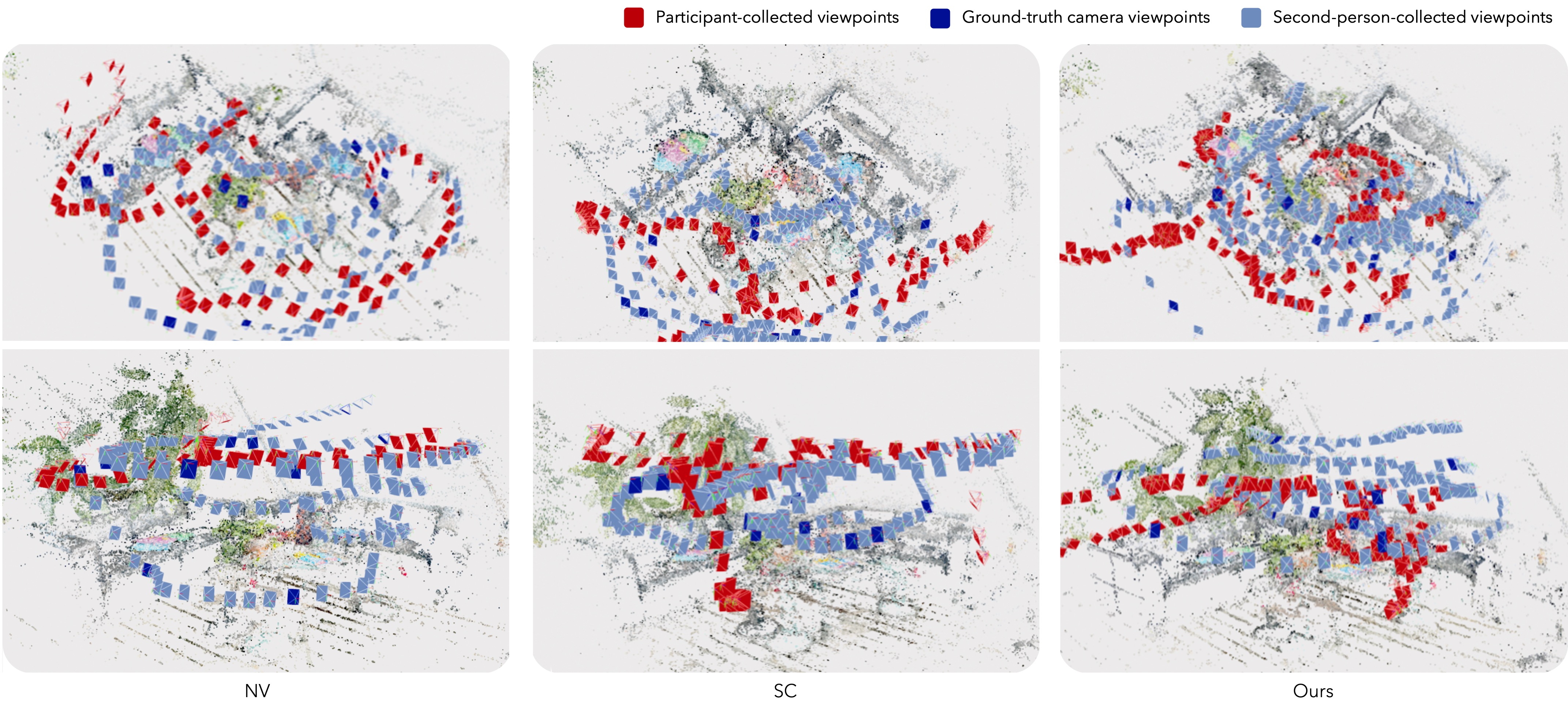}
    \caption{Camera viewpoint visualization in the Lounge scene with \texttt{NV}, \texttt{SC}, and \texttt{Ours}. 
    Red: Training viewpoints collected by the participants; Blue: Ground-truth viewpoints randomly sampled from images collected by a second person to avoid arbitrary evaluation; Light blue: All the other viewpoints collected by a second person.}
    \label{fig:view_distributions}
\end{figure*}

\begin{figure}[t]
    \centering
    \includegraphics[width=\columnwidth]{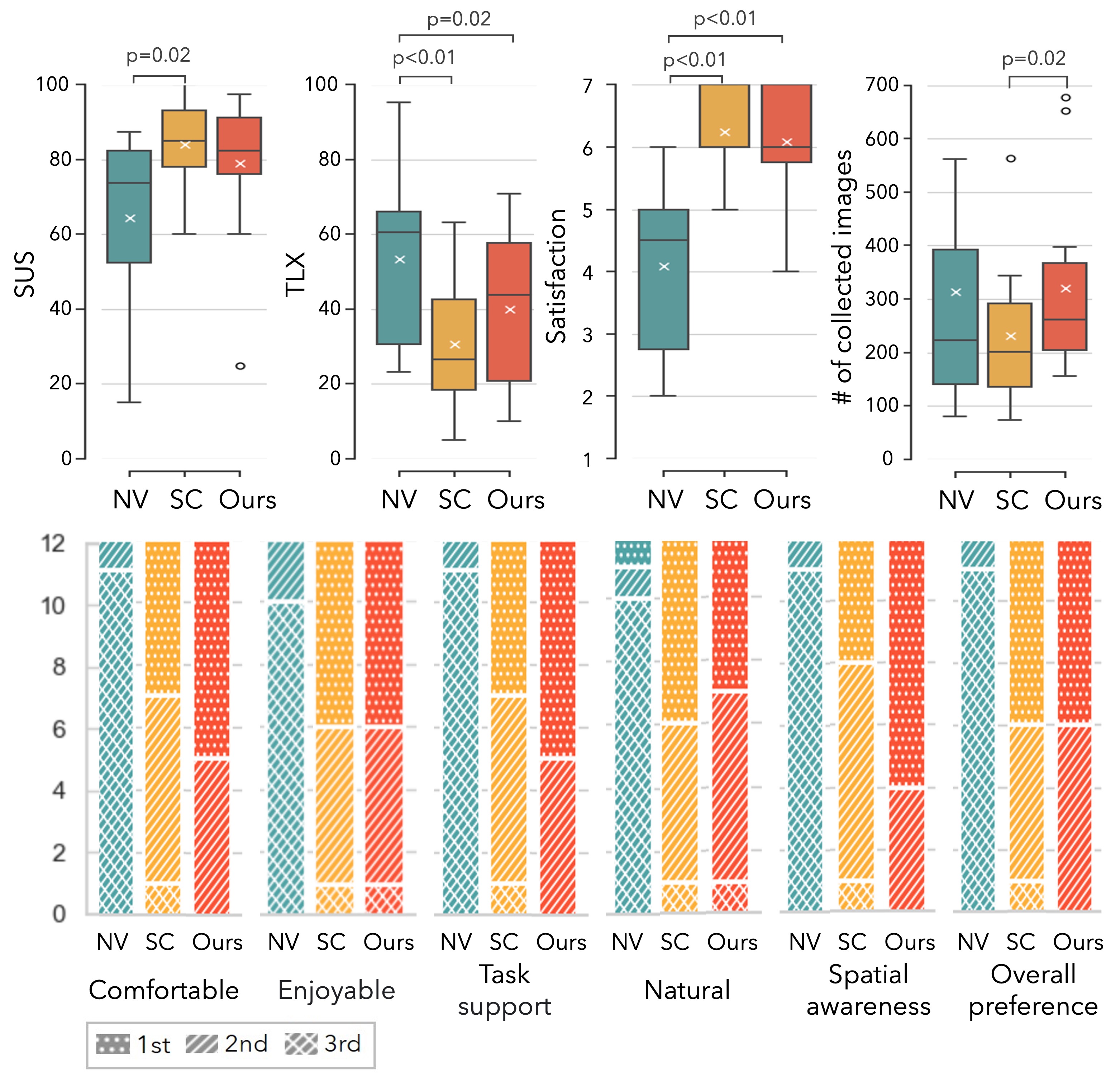}
    \caption{Results of the user study.
    }
    \label{fig:userstudy_results}
\end{figure}

\subsection{User preferences and comments}

\paragraph{Results}
We performed either ANOVA or Friedmann tests depending on whether the data sphericity and normality were met.

The Friedman test revealed a significant main effect on SUS ($\chi^2(2)=10.30$, $p=0.006$, Kendall's $W=0.43$). Post-hoc pairwise comparisons using Wilcoxon signed-rank tests with Bonferroni correction indicated a significant difference (\texttt{NV}: $\Bar{X}=64.38$, SD $=22.41$; \texttt{SC}: $\Bar{X}=83.96$, SD $=10.97$; $p=0.02$, $r=0.74$).
The ANOVA revealed a significant main effect on TLX (F$(2,22)=9.88$, $p<0.001$, $\eta^2=0.17$). Post-hoc pairwise comparisons (Benjamini-Hochberg FDR correction) indicated significant differences between \texttt{NV} and \texttt{SC} (\texttt{NV}: $\Bar{X}=53.42$, SD $=21.17$; \texttt{SC}: $\Bar{X}=30.53$, SD $=18.77$; $p<0.01$, Cohen's d $=1.10$) and between \texttt{NV} and \texttt{Ours} (\texttt{NV}: $\Bar{X}=53.42$, SD $=21.17$; \texttt{Ours}: $\Bar{X}=39.92$, SD $=21.34$; $p=0.02$, Cohen's d $=0.61$).

The Friedman test revealed a significant main effect on Satis ($\chi^2(2)=15.20$, $p=0.0005$, Kendall's $W=0.63$).
Post-hoc pairwise comparisons indicated significant differences between \texttt{NV} and \texttt{SC} (\texttt{NV}: $\Bar{X}=4.08$, SD $=1.44$; \texttt{SC}: $\Bar{X}=6.25$, SD $=0.72$; $p<0.01$, $r=0.83$) and also between \texttt{NV} and \texttt{Ours} (\texttt{NV}: $\Bar{X}=6.25$, SD $=0.72$; \texttt{Ours}: $\Bar{X}=6.08$, SD $=0.95$; $p<0.01$, $r=0.83$).
The Friedman test revealed a significant main effect on the number of collected images ($\chi^2(2)=6.50$, $p=0.04$, Kendall's $W=0.27$). Post-hoc pairwise comparisons indicated significant difference (\texttt{SC}: $\Bar{X}=231.67$, SD $=131.31$; \texttt{Ours}: $\Bar{X}=320.17$, SD $=169.03$; $p=0.02$, $r=0.75$).

Although several effects reached statistical significance, some effect sizes (e.g., Kendall’s $W < 0.70$) indicated only moderate agreement.
Given the limited sample size, the statistical power for detecting such effects may be insufficient. This limitation should be considered when interpreting the results, and future studies with larger samples are warranted to confirm these findings.

\figurename~\ref{fig:userstudy_results}-bottom presents the results of the post-task questionnaires. No participants selected \texttt{NV} as their top choice for any question. Between \texttt{Ours} and \texttt{SC}, a noticeable difference emerged for Q5, with \texttt{Ours} providing more effective three-dimensional spatial feedback. \texttt{SC} was ranked lowest in all questionnaires once, while \texttt{Ours} was ranked lowest twice, in Q2 and Q4. Otherwise, no critical differences were observed between the two.

\begin{table}[t]
    \centering
    \caption{Quantitative comparisons against baseline data collection strategies in view synthesis quality. }
    \label{table:results}
    \footnotesize
    \begin{tabular}{lccc}
        \multicolumn{4}{c}{3DGS \cite{Kerbl20233DGS}} \\
        \toprule
         & PSNR ($\uparrow$) & SSIM ($\uparrow$) & LPIPS ($\downarrow$) \\
        \midrule
        \texttt{NV} & 16.338 (3.346) & 0.752 (0.142) & 0.292 (0.120) \\
        \texttt{SC} & 17.086 (3.435) & 0.800 (0.132) & 0.253 (0.112) \\
        \midrule
        \texttt{Ours} & \textbf{18.891 (3.195)} & \textbf{0.848 (0.111)} & \textbf{0.201 (0.102)} \\
        \bottomrule
    \end{tabular}
    \begin{tabular}{lccc}
        \vspace{-0.5em}\\
        \multicolumn{4}{c}{Nerfacto \cite{nerfstudio} } \\
        \toprule
         & PSNR ($\uparrow$) & SSIM ($\uparrow$) & LPIPS ($\downarrow$) \\
        \midrule
        \texttt{NV} & 16.115 (2.592) & 0.558 (0.088) & 0.589 (0.072) \\
        \texttt{SC} & 16.372 (2.488) & 0.574 (0.075) & 0.570 (0.069) \\
        \midrule
        \texttt{Ours} & \textbf{18.134 (1.859)} & \textbf{0.616 (0.068)} & \textbf{0.520 (0.065)} \\
        \bottomrule
    \end{tabular}
\end{table}

\paragraph{Discussion}
\texttt{SC} received higher SUS scores compared to \texttt{NV}. No statistically significant difference was observed for \texttt{Ours} in SUS. However,
the system performance of \texttt{Ours} was well acknowledged in the post-trial evaluations (\figurename~\ref{fig:userstudy_results}-bottom).
We believe that the additional spherical proxies introduced in \texttt{Ours} may have contributed to an increased cognitive load for participants, which could explain the slightly lower SUS scores.

Similar trends were observed in NASA-TLX and Satis scores. Both \texttt{Ours} and \texttt{SC} showed significantly lower mental workload and higher satisfaction compared to \texttt{NV}. No significant difference was found between \texttt{Ours} and \texttt{SC}. Since \texttt{NV} lacked any visual guidance, a participant expressed uncertainty or insecurity about whether all parts of the scene had been sufficiently scanned (P8).

Participants using \texttt{Ours} captured more images than those using \texttt{SC}, which may account for the increased perceived workload.
Several participants noted challenges related to angular sampling, such as ``Getting all the suggested angles is cumbersome'' (P5, P6, P8) and ``There are some angles that are difficult to reach'' (P10).
Interestingly, participants using \texttt{NV} and \texttt{Ours} collected a similar number of images, while \texttt{SC} resulted in significantly fewer captures. 
The fact that \texttt{Ours} achieved the best view synthesis performance (\tablename~\ref{table:results}) suggests that it guided users toward additional viewpoints \texttt{SC} failed to capture. It also demonstrates a more efficient view sampling than \texttt{NV}.
Some participants noted that they would have better understood and appreciated the additional effort if they had seen the final rendered results. Four participants (P1, P2, P8, P9) specifically mentioned this, although real-time feedback was not possible due to the time-consuming nature of COLMAP and 3DGS (or Nerfacto) optimization, which can take several hours.

Participants also gave positive feedback on \texttt{Ours}, such as ``I had fun making the spheres disappear'' (P3), and ``\texttt{Ours} allows me to estimate how many angles I got and how good the result may look'' (P8).
As for feature requests, the most frequently requested improvement was the ability to preview the final rendered result immediately. The second most requested feature was a clearer signal for task completion.
Due to the progressive nature of \texttt{Ours}, the participants were free to explore anywhere, while it was roughly restricted to the areas shown in \figurename~\ref{fig:exp_setup}.
In some parts, the participants were frustrated by unreachable areas (P7, P12).

\figurename~\ref{fig:view_distributions} visualizes typical data of the participants using different approaches. Since the participants only saw the video stream in \texttt{NV}, they had to rely on their intuition to navigate through the environment. As some participants pointed out, they had no clear clues about what they had captured. Thus, the viewpoints appear distributed rather randomly.
\texttt{SC} visualizes which areas have already been observed, and the overall viewpoints appear to cover the space more evenly than in \texttt{NV}. However, the viewpoints do not travel in three dimensions, instead staying at the same height.
\texttt{Ours} allowed the participants to capture the necessary number of viewpoints depending on the complexity of the 3D structure of the scene. As visual proof, the viewpoints traveled comprehensively within the space, while partly focusing on a cluster of objects on the table.

We calculated the distances between views collected by the participants and those of the ground truth. Since the scale between point clouds was unknown, we employed Coherent Point Drift (CPD) \cite{myronenko2010point} to match the scales between different structure-from-motion point clouds.
We used a known size of a glass bottle to calculate the real-world distances on a metric scale.
The average distances to the ground truth viewpoints (mean $\pm$ SD) for \texttt{NV}, \texttt{SC}, and \texttt{Ours} were \num{0.337} \ensuremath{\pm} \num{0.173}~\si{\meter}, \num{0.320} \ensuremath{\pm} \num{0.155}~\si{\meter}
, and \num{0.239} \ensuremath{\pm} \num{0.047}~\si{\meter}, respectively, with \texttt{Ours} achieving the smallest distance.
The average angular differences to the ground truth viewpoints for \texttt{NV}, \texttt{SC}, and \texttt{Ours} were \num{17.060} $\pm$ \num{8.189}~\si{\degree}, \num{18.540} $\pm$ \num{5.405}~\si{\degree}, and \num{12.392} $\pm$ \num{4.338}~\si{\degree}, with \texttt{Ours} also achieving the smallest angular difference.
The ground truth viewpoints were designed to be diverse by randomly sampling camera positions and orientations. Therefore, the fact that \texttt{Ours} exhibited the smallest distance and angular difference to the ground truth suggests that our method effectively captures the scene from a diverse range of viewpoints without bias or missing coverage.

\section{Limitations and Future Directions}

While IntelliCap demonstrated strong performance in efficient view sampling, it also revealed several unique limitations and opened up interesting directions for future research.

\paragraph{Level of detail control}
We do not fully utilize the LLM scores. One possible extension would be to adaptively increase the number of sphere proxy subsurfaces based on the scores, providing finer granularity for high-priority objects.

\paragraph{Open vocabulary}
Our system relies on a closed vocabulary system for lightweight mobile implementation used in the user studies.
For more scene-adaptive content validation, future work could incorporate more intelligence into this view sampling task, such as vision-language models or an open vocabulary system, to achieve higher flexibility and recognition performance.

\paragraph{Full mobile implementation}
We rely on a client-server model to offload vision processing from a mobile device. The most computationally intensive part of our pipeline is Detectron2, which can technically be implemented on mobile devices.
We further anticipate that mobile neural processors will bring more mobile-friendly vision processing to enhance view-sampling tasks.

\paragraph{Wide FoV imaging and displaying}
Due to video stabilization, the current generation of smartphones decreases FoV when AR functionalities are active.
Despite this, a larger FoV is advantageous to our task as it allows faster spatial scanning and object identification for both machines and human operators.
However, note that different FoV in imaging and displaying would require additional adjustments to our visualization approaches.

\paragraph{Large scale and outdoor scenes}
While our approach can technically be scaled to larger scenes, our current results are limited to below 10 m$^2$ to perform the user study in a reasonable time, as well as additional immediate view sampling of the ground truth dataset to conduct view synthesis evaluations.
Here, further development in evaluation schemes for multiple users is needed.
Applying the method to outdoor scenes would face more severe temporal and weather changes that can lead to inconsistent brightness, shadows, and noises.
Dynamics would be handled by identifying and excluding potentially moving objects such as \texttt{person}, \texttt{car}, and \texttt{bicycle} \cite{yang2024enhancing} or using sophisticated reconstruction methods \cite{Schischka2025DynaMon}.
Precise sphere alignment becomes increasingly critical for distant objects, while the generation can be constrained to reachable areas.

\paragraph{Efficient task completion}
We concluded that potential increases in time and mental workload from additional visualizations and longer photographing sessions would be acceptable to obtain better final view synthesis results.
To improve efficiency and user engagement during the view sampling process, multimodal feedback such as visual indicators, audio cues, and haptic responses can help guide and motivate users.
These enhancements might include progress bars and rewarding feedback like confirmation sounds or vibrations upon view sampling.
However, as discussed in Section \ref{sec:related_work}, defining ``sufficient'' sampling and task ``completion'' remains inherently difficult, as these can only be properly evaluated after the final view synthesis.

\paragraph{Misidentification and misalignment}
Several factors can hinder the effectiveness of our approach. For instance, Detectron2 may occasionally misclassify a simple object as complex or vice versa. Although we did not observe such cases during our study, such misclassifications could lead users to capture an excessive number of images unnecessarily or overlook important objects.
Sphere misalignment is another potential issue, which may result in floating spheres that confuse users during the capture process.
Our sphere merging strategy is designed to address multiple detections of cluttered, complex objects by grouping them efficiently, thereby enabling the capture process to be completed within a reasonable timeframe. However, in edge cases where the merging policy reaches the maximum allowable sphere size, a smaller adjacent sphere may be generated. This can lead to suboptimal grouping and reduced efficiency in certain scenarios.

\paragraph{Diversity}
Our visualization approach is grounded in heuristically established techniques and has been positively acknowledged throughout the study. However, there remains significant potential to explore more inclusive methods that accommodate individuals with limited color vision, hearing impairments, mobility challenges, and a variety of smartphone devices in their pockets.
We believe this represents a new venue for visualization, human-computer interaction, and accessibility research.
Furthermore, when moving at excessively high speeds, motion blur can occur in captured images. To mitigate this motion blur, we believe that measuring scan speed and considering user variability would be effective.

\section{Conclusion}

This paper presents a novel view sampling approach for high-quality view synthesis, capable of handling multiple scales and complex scenes. Using object semantic classification and LLM to assess the scanning requirements of observed objects, our system enables users to discover more diverse and informative viewpoints, resulting in improved synthesis quality.
Unlike conventional best practices that prioritize spatial or angular coverage, our method combines both to enable progressive scanning. The system predicts regions and objects that require denser sampling to support human operators without prior knowledge of scene contents and view synthesis. Experimental results demonstrate that this strategy leads to consistent rendering performance and effective task execution.

Given the novelty of our task design, we introduced new evaluation schemes to assess both the view sampling process and the resulting synthesis quality. Specifically, we designed a user study to collect subjective feedback on our visualization choices by comparing different sampling strategies and developed a follow-up sampling scheme performed by examiners to evaluate view synthesis performance. We plan to release our source code to support future research in this emerging area, particularly contributions aimed at enhancing visual feedback for strategic view sampling coverage and user interaction.

\acknowledgments{
 This work was supported by 
 the Alexander von Humboldt Foundation funded by the German Federal Ministry of Education and Research,
 the Deutsche Forschungsgemeinschaft (DFG, German Research Foundation) under Germany's Excellence Strategy – EXC 2120/1 – 390831618,
 and partly by a grant from JST Support for Pioneering Research Initiated by the Next Generation (\# JPMJSP2123)
}

\bibliographystyle{abbrv-doi}

\bibliography{references}

\end{document}


\firstsection{Introduction}

\maketitle

\begin{figure*}[!t]
    \centering
    \includegraphics[width=\textwidth]{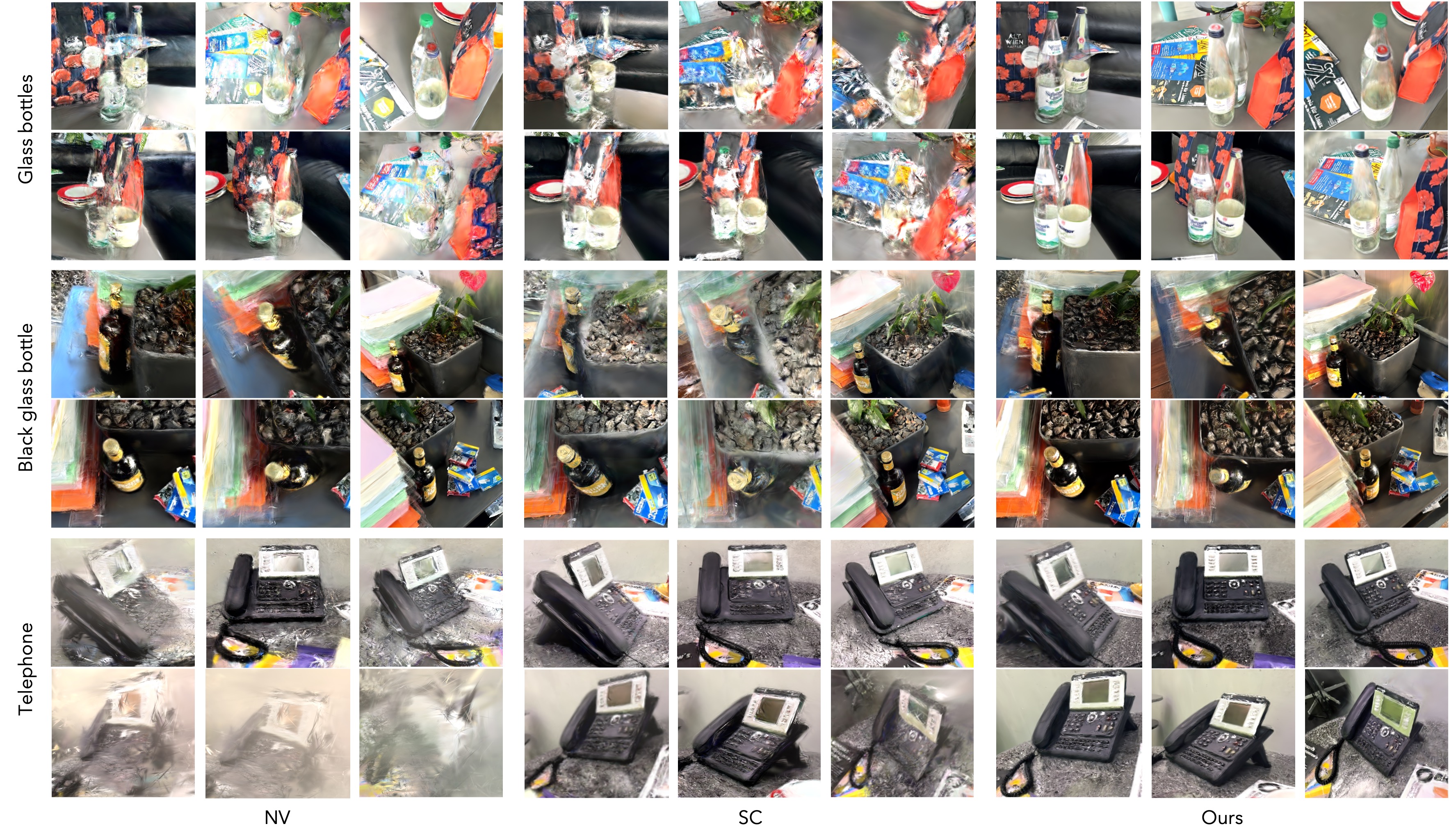}
    \caption{3DGS \cite{Kerbl20233DGS} view synthesis results using data collected by ours and baseline approaches, showing how reflective and transparent objects are reproduced at different angles from the collected views.}
    \label{fig:obj_center_3dgs}
\end{figure*}

This supplemental document presents view synthesis results using data collected by our method and the baseline approaches, demonstrating how reflective and transparent objects are reproduced from different angles.

Please also refer to the supplemental video that demonstrates our IntelliCap and baseline approaches, and their corresponding view synthesis results in motion.

\begin{figure*}[!t]
\vspace{4.2mm}
    \centering
    \includegraphics[width=\textwidth]{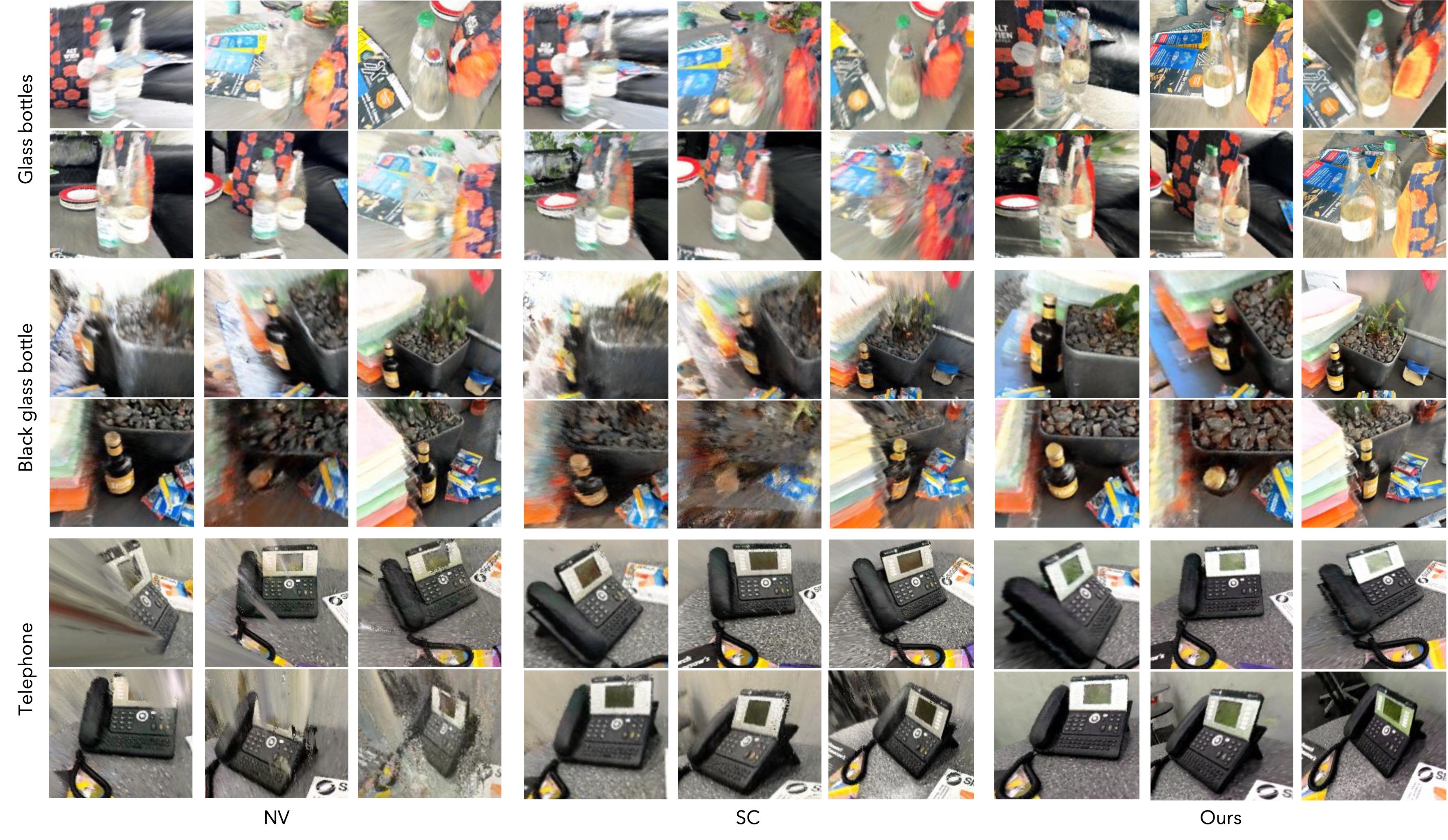}
    \caption{Nerfacto \cite{nerfstudio} view synthesis results using data collected by ours and baseline approaches, showing how reflective and transparent objects are reproduced at different angles from the collected views.}
    \label{obj_center_nerfacto}
\end{figure*}

\acknowledgments{
 This work was supported by 
 the Alexander von Humboldt Foundation funded by the German Federal Ministry of Education and Research,
 the Deutsche Forschungsgemeinschaft (DFG, German Research Foundation) under Germany's Excellence Strategy – EXC 2120/1 – 390831618,
 and partly by a grant from JST Support for Pioneering Research Initiated by the Next Generation (\# JPMJSP2123)
}

\bibliographystyle{abbrv-doi}

\bibliography{references}